\documentclass[lettersize,journal]{IEEEtran}

\usepackage{amsmath,amsfonts}
\usepackage{algorithmic}
\usepackage{algorithm}
\usepackage{array}
\usepackage[caption=false,font=normalsize,labelfont=sf,textfont=sf]{subfig}
\usepackage{textcomp}
\usepackage{stfloats}
\usepackage{url}
\usepackage{verbatim}
\usepackage{graphicx}
\usepackage{cite}
\usepackage{orcidlink}
\usepackage[utf8]{inputenc}
\usepackage[small]{caption}
\usepackage{graphicx}
\usepackage{amsmath}
\usepackage{amsthm}
\usepackage{amsfonts} 
\usepackage{booktabs}
\usepackage{algorithm}
\usepackage{algorithmic}
\usepackage[switch]{lineno}
\usepackage{xcolor}
\usepackage{enumitem}
\usepackage{subfig}
\usepackage{svg}
\usepackage{multirow}
\usepackage{xcolor,colortbl}
\definecolor{MLP}{HTML}{CBA6D1}
\definecolor{RF}{HTML}{FFBF80}
\definecolor{Seq}{HTML}{F18C8D}
\definecolor{KW}{HTML}{D3AA93}
\definecolor{Simple}{HTML}{A6D7A4}
\definecolor{LLM}{HTML}{5BB7DD}
\usepackage{lscape}
\usepackage{alphabeta}

\hypersetup{
    colorlinks,
    linkcolor={black},
    citecolor={black},
    urlcolor={black}
}

\newtheorem{definition}{Definition}

\begin{document}

\title{Agnostic Visual Recommendation Systems: Open Challenges and Future Directions}
\author{Luca Podo \orcidlink{0000-0001-8780-6848}, Bardh Prenkaj \orcidlink{0000-0002-2991-2279
}, Paola Velardi \orcidlink{0000-0003-0884-1499}
        % <-this % stops a space
\thanks{All authors are with the Computer Science Department of Sapienza University of Rome, Via Salaria 113, Rome, Italy (emails: [podo,prenkaj,velardi]@di.uniroma1.it}% <-this % stops a space
%\thanks{Manuscript received April 19, 2021; revised August 16, 2021.}}
}

\markboth{Journal of \LaTeX\ Class Files,~Vol.~14, No.~8, August~2021}%
{Shell \MakeLowercase{\textit{et al.}}: A Sample Article Using IEEEtran.cls for IEEE Journals}


\twocolumn[{%
\renewcommand\twocolumn[1][]{#1}%
\maketitle
\begin{center}
    \centering
    \captionsetup{type=figure}
    \includegraphics[width=.9\textwidth]{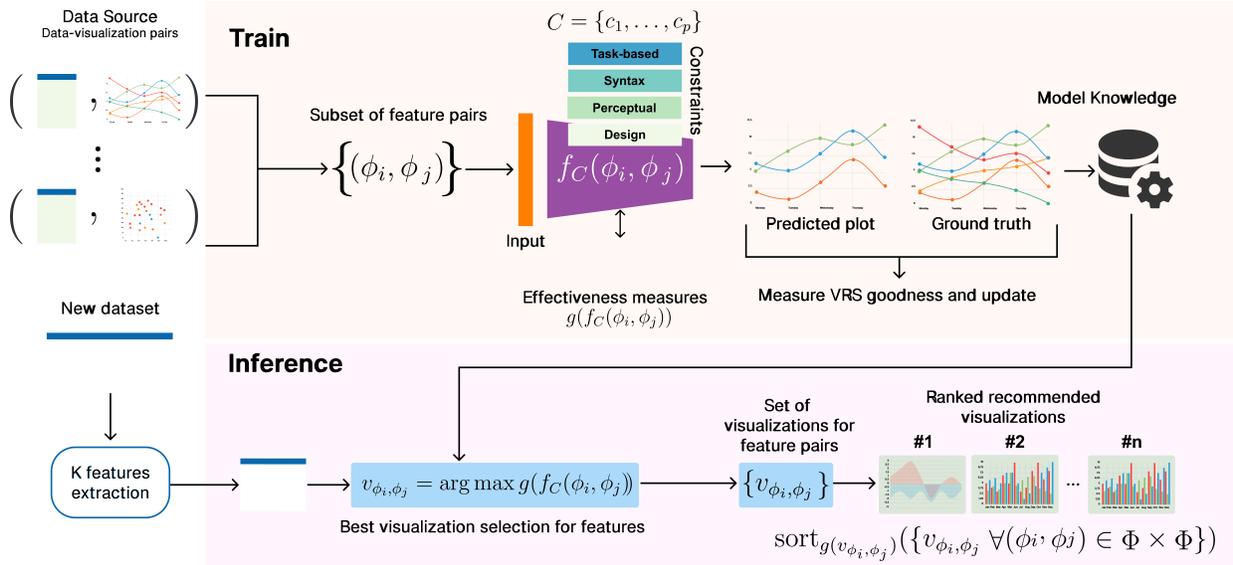} 
    \captionof{figure}{Workflow of Agnostic Visual Recommender Systems (A-VRSs).}
    \label{fig:workflow}
\end{center}%
}]

\begin{abstract}
Visualization Recommendation Systems (VRSs) are a novel and challenging field of study aiming to help generate insightful visualizations from data and support non-expert users in information discovery. Among the many contributions proposed in this area, some systems embrace the ambitious objective of imitating human analysts to identify relevant relationships in data and make appropriate design choices to represent these relationships with insightful charts.  We denote these systems as ``agnostic'' VRSs since they do not rely on human-provided constraints and rules but try to learn the task autonomously. Despite the high application potential of agnostic VRSs, their progress is hindered by several obstacles, including the absence of standardized datasets to train recommendation algorithms, the difficulty of learning design rules, and defining quantitative criteria for evaluating the perceptual effectiveness of generated plots. This paper summarizes the literature on agnostic VRSs and outlines promising future research directions.  
\end{abstract}


\section{Introduction}
In recent years, data-driven approaches have gained increasing significance across various sectors of society and science, fundamentally transforming how strategies and decisions are made. The vast availability of data represents a remarkable opportunity, enabling organizations to gain previously unattainable insights. However, this data abundance also brings with it complexities and challenges related to data interpretation and utilization \cite{chin2019three,rudin2022interpretable,zhang2021survey,peake2018explanation,chen2022grease}.

Data acquisition, analysis, and transformation into actionable insights have given rise to a substantial demand for data scientists. These professionals have the skills and knowledge to navigate the data landscape, uncover meaningful patterns, and inform decision-making processes. Nevertheless, the supply of these specialized data scientists often falls short of meeting the burgeoning needs of the industry\footnote{\url{https://www.techtarget.com/searchbusinessanalytics/feature/Data-scientist-shortage-leaves-organizations-uncertain}}.

This shortage of data scientists creates a significant gap between the data-rich environment and the capacity of non-expert users to harness the potential of data. In this context, Visualization Recommendation Systems (VRSs) come into play, as also discussed in \cite{saket2018beyond}. A VRS is designed to bridge this gap by providing automated tools that facilitate visual discovery, making data analysis more accessible to a wider audience across diverse sectors and applications.

For high-dimensional data, in particular, a VRS is a valuable asset for data analysts. They assist in narrowing down the vast search space of potential insightful visualizations, allowing analysts to focus on the most relevant and informative representations of the data\cite{wang2021survey,zhu2020survey}. This enhances the efficiency of data analysis and empowers data professionals to make more informed decisions \cite{graf2020supporting,mottus2015use,shen2002data,oghbaie2016understanding}.

More recently, VRSs have faced a new challenge: producing insightful visualizations based solely on the datasets provided as input, without human-provided constraints and rules. Hereafter, we denote these types of VRS as Agnostic Visual Recommender Systems (A-VRSs) since they are “not knowing” (ἄγνωστος) a priori which variables in the input dataset have relevant relationships, nor how they should be rendered at best in a plot. VRSs pursue the ambitious goal of learning to mimic human analysts to extract relevant relationships from data and present them with insightful charts. In this regard, A-VRS can be considered to embrace the objective of so-called Artificial General Intelligence (AGI) \cite{AGI}. Their role is pivotal in Machine Learning for Visual Analytics (ML4VIS), which harnesses machine learning techniques to improve visualizations' design, development, and evaluation. The fusion of AI and data visualization enables the creation of intelligent systems that aid in interpreting and communicating complex data \cite{wu2021ai4vis}.

As we navigate the era of big data, A-VRSs hold immense potential for enhancing decision-making processes by helping individuals, both experts and non-experts, gain valuable insights from data. However, despite their promise, numerous challenges slow down the rapid progress of A-VRS. These challenges span various aspects, from data representation to recommendation algorithms and evaluation. This survey, which is -- to the best of our knowledge -- the first in this area,  seeks not only to assess the state of the art of A-VRS and VRSs in general but also to identify the obstacles limiting their diffusion and to outline possible new promising research opportunities. 
\section{Method to conduct this Survey and Contributions}

In this section, first, we highlight the relevance and potential of the VRS field (Sec. \ref{significance}), which justifies the need for a survey study. Next  (Sec. \ref{surveymethodology}), we describe the methodology to collect relevant papers and compare our work with other surveys in the broader domain of ML4VIS. Finally  (Sec. \ref{contribution}), we summarize the contributions of our work.
\subsection{The Significance of VRSs}
\label{significance}

VRSs represent a pivotal and rapidly evolving field at the intersection of machine learning and data visualization. Their growing importance is underpinned by several compelling motivations, each of which underscores the critical role VRSs play in addressing contemporary challenges and driving innovation. Here, we elucidate why future researchers should care about VRSs (particularly A-VRSs ) and their broader implications in various domains.

\noindent\textbf{Enhancing Data Understanding and Analytical Reasoning}.
VRSs offer a powerful solution to modern society's increasing complexity and volume of data. With the proliferation of data sources and types, such as sensor data, social media content, and scientific datasets, the ability to effectively comprehend and access relevant information is paramount~\cite{ellen2014barriers}. VRSs act as intelligent intermediaries that bridge the gap between users and complex data, making it more accessible and understandable. By providing personalized visual recommendations, VRSs empower individuals, researchers, and decision-makers to extract meaningful insights from data, thus facilitating informed decision-making~\cite{packer2018visually,long2020personalized,livne2022evolving}.

\noindent\textbf{Optimizing Decision Support and Expertise}.
VRSs have substantial implications in fields where data-driven decisions and domain expertise are critical, including healthcare~\cite{electronics12173709,valdez2016recsys4healthinformatics}, finance~\cite{sharaf2022recsys4finance}, and scientific research~\cite{liang2021effect}. By leveraging machine learning techniques, VRSs can guide domain experts, medical practitioners, and financial analysts to explore and analyze data that complements their expertise. This, in turn, leads to more informed and accurate decision-making, cost savings, and even life-saving interventions~\cite{rasmussen2010nimble,franco2019mentor}. In scientific research, VRSs can assist in rapidly comprehending complex datasets, accelerating the pace of discovery and innovation~\cite{vartak2016towards,alofabi019recsys4health}.

\noindent\textbf{Meeting Accessibility and Inclusivity Needs}.
VRSs, particularly A-VRSs, can make data and information more accessible to individuals with varying degrees of visual and cognitive abilities \cite{zhang2019deep}. By tailoring visual representations to accommodate specific needs, VRSs can contribute to greater inclusivity and accessibility~\cite{joyner2022visualization} in various domains, including digital content consumption, user interfaces, and educational materials. 

The significance of VRSs extends far beyond a mere convenience in data visualization. Their potential impact spans fields as diverse as data science~\cite{suh2023metrics,vartak2015seedb}, healthcare~\cite{demiralp2017foresight,spicer2022creating}, education~\cite{evagorou2015role,spicer2022creating}, and creative design~\cite{nguyen2021examining}. As VRSs evolve towards A-VRSs, their ability to facilitate data understanding, enhance user experiences, optimize decision support, and promote inclusivity underscores their critical role in addressing contemporary challenges and driving innovation in a data-rich world.

\subsection{Methodology and comparison with other surveys}
\label{surveymethodology}

A-VRSs are still in their early stages since significant attention to machine learning for visualization purposes has risen only in the past few years. This work aims to survey existing studies addressing VRSs in general, focusing on the emerging field of A-VRSs, providing an in-depth discussion and a systematic literature review. To achieve this goal, we conducted a secondary mapping study, according to the definition and method provided in~\cite{surveymethod}  and~\cite{findingkeywords}: a primary study is an empirical investigation into specific research questions, while a secondary study is a review of primary studies related to specific research questions to synthesize the results. 
 
To identify the keywords of the search string for retrieving relevant works, we followed the general approach suggested in ~\cite{findingkeywords}. Given the relatively narrow domain, we started with the following focused search string:
\begin{flalign*}
      & \text{\textit{ML4VIS} \texttt{OR} \textit{ML2VIS} \texttt{OR} \textit{VRS} \texttt{OR} \textit{AI4VIS}}\\
      & \text{\texttt{OR} (\textit{''artificial intelligence}" \texttt{AND} \textit{visualization})}\\
      & 
      \text{\texttt{OR} \textit{''automatic visualization generation"}}\\
      & \text{\texttt{OR} ((\textit{visualization} \texttt{OR} \textit{visual}) \texttt{AND} \textit{recommender})}
\end{flalign*}

\noindent After analyzing the first set of retrieved papers, we extended the search with 
\begin{flalign*}
      & 
\text{\textit{AUTOVIS} \texttt{OR}}\\
     & 
\text{((\textit{LLM2VIS} \texttt{OR} \textit{''large language models"})}\\ & \text{\texttt{AND} \textit{''visualization"})}
\end{flalign*}

Next, we organized the retrieved  VRSs according to a uniform formal notation to facilitate an easy comparison of their strengths and pitfalls (see Sec. \ref{sec:problem_formulation} and \ref{sec:methods}). Although, for completeness, we shortly survey relevant VRSs, including rule-based and task-aware ones, our focus is on A-VRSs. We propose a formalization of A-VRSs when considering the learning of visualization constraints of different types (see Sec. \ref{sec:vizconstraints}). We summarize the strengths and weaknesses of A-VRSs in the literature, classifying them according to several dimensions, making it easy for readers to identify alternative methodologies that better suit their scenario (see Sec. \ref{sec:related_work}). Sec. \ref{sec:open_challenges} highlights unresolved issues in this area of research. Sec. \ref{sec:future_work} suggests potential future avenues for addressing the previously mentioned challenges. Finally, Sec. \ref{sec:conclusion} concludes the paper.

To the best of our knowledge, this survey is the first to tackle VRSs comprehensively and to highlight the potential of A-VRSs. However, we analyzed surveys that adopt a more general scope for completeness purposes, i.e., ML/AI for visualization~\cite{wang2021survey,wu2021ai4vis}. We argue that our survey has a narrower scope than previous ones, and it delves much deeper into the formal notation and the analysis of proposed methods, benchmarking datasets, and measures of visualization effectiveness, as summarized in Table \ref{tab:survey_comparison}. 
\begin{table}[!t]
\centering
\caption{Qualitative comparison of  Intelligent VRS in the literature, according to the dimensions reflecting the scope and workflow of VRS. $\cdot$ depicts a missing aspect, \checkmark a covered aspect, and $\sim$ a partially covered aspect.}
\label{tab:survey_comparison}
\resizebox{\linewidth}{!}{%
\begin{tabular}{@{}llccc@{}}
\toprule
\multicolumn{2}{l}{} &
  [this] &
  Wang et al. \cite{wang2021survey} &
  Wu et al. \cite{wu2021ai4vis} \\ \midrule
\multirow{3}{*}{Domain}          & VRS                     & \checkmark & $\cdot$    & $\cdot$    \\
                                 & ML4VIS                  & $\cdot$    & \checkmark & \checkmark \\
                                 & AI4VIS                  & $\cdot$    & $\cdot$    & \checkmark \\ \midrule
Definition                       &                         & \checkmark & \checkmark & $\cdot$    \\ \midrule
\multirow{3}{*}{\begin{tabular}[c]{@{}l@{}}Visualization\\ Generation\end{tabular}} &
  Constraints Formalization &
  \checkmark &
  $\cdot$ &
  $\cdot$ \\
                                 & Effectiveness Measures  & \checkmark & $\cdot$    & $\cdot$    \\
                                 & Visual Interpretability & \checkmark & $\cdot$    & $\cdot$    \\ \midrule
\multirow{3}{*}{Method Taxonomy} & Rule-based              & \checkmark & $\cdot$    & $\cdot$    \\
                                 & ML-based + LLMs         & \checkmark & $\sim$     & $\sim$     \\
                                 & Hybrid-based            & \checkmark & \checkmark & \checkmark \\ \midrule
\multirow{3}{*}{Datasets}        & Crowdsourcing           & \checkmark & $\cdot$    & $\cdot$    \\
                                 & Rule-based              & \checkmark & $\cdot$    & $\cdot$    \\
                                 & Custom-generated        & \checkmark & $\cdot$    & $\cdot$    \\ \midrule
Visualization Pipeline           &                         & $\sim$     & \checkmark & $\sim$     \\ \bottomrule
\end{tabular}%
}
\end{table}
We analyzed these existing surveys according to six macro-dimensions, reflecting the scope, the domains, and the typical task workflow considered in the surveys. The first dimension, \textit{Domain}, considers the breadth and scope of each survey. \textit{Definition} refers to providing formal and uniform definitions for the considered domain. \textit{Visualization Generation} considers the strategy or measures to provide the end-user with the most relevant and effective visualizations. \textit{Method's Taxonomy} refers to categorizing the surveyed works in the literature. \textit{Datasets} is about enlisting different types of data-visualization pairs in the wild data collections. \textit{Visualization Pipeline} presents a step-by-step guide on how the user interacts with the VRSs to engender specific visualizations. 

Every dimension presented in Table \ref{tab:survey_comparison} is tailored to the specific domain where the surveyed works are situated to conduct an equitable comparison. For instance, in the \textit{Visualization Pipeline} dimension, our survey delineates the procedural workflow of VRSs while, when examining the same dimension in the survey by Wang et al. \cite{wang2021survey}, we scrutinize whether their work outlines a systematic process of integrating Machine Learning models for visualization purposes.

\noindent\textbf{Domain} -- We identified three main scope classes: i.e., VRS, ML4VIS, and AI4VIS. The other surveys concentrate on the broader domain of AI4VIS and offer a general overview and scaffolding for the application of AI in the data visualization field. Wang et al.~\cite{wang2021survey} propose an exhaustive overview of the visualization tasks that could potentially benefit from the application of machine learning techniques (e..g, style transfer or visualization interaction). Their survey addresses the question \textit{"Which aspects of the visualization process can benefit from the application of Machine Learning?} Similarly, Wu et al.~\cite{wu2021ai4vis} discuss all the possible opportunities to involve artificial intelligence approaches to support data visualization. Using the methodology of \textit{What}, \textit{Why}, and \textit{How}, they initially discuss the data types that can be integrated into this task, such as programs, graphics, or hybrid. Then, they identify three categories of applications that fit why AI can play a significant role in visualization. Lastly, they illustrate how AI can be effectively integrated by providing different use cases and reviewing the literature for this particular task. While these surveys encompass a broader range of intelligent systems to support visualization, they are not concerned with the specific topic of VRSs from the perspective of applying visualization generation constraints. We bridge this gap in the literature and provide the reader with a survey of recent and novel works.

\noindent\textbf{Definition} -- Although AI4VIS, and the less broad ML4VIS, have been treated extensively (among many~\cite{luo2018deepeye,hu2019vizml,dibia2019data2vis,wu2021multivision,dibia2023lida}), this survey is the first contribution in the literature which provides a formal definition of A-VRSs. Our goal is to provide future researchers with a principled definition that aids them in aligning their work with the SoTA.
Note that in Table \ref{tab:survey_comparison}, this dimension is reported as missing for the other two surveys since they do not provide a formal definition concerning their specific analysis domain.

\noindent\textbf{Visualization Generation} -- We argue that incorporating constraints is necessary when generating informative visualizations (see Sec. \ref{sec:vizconstraints}), an aspect overlooked by previous surveys. These constraints put the space of all possible visualizations into a funnel that sieves through those that have no meaning for a particular use case. To this end, we define four classes of constraints, each affecting a specific component of the engendered visualizations (e.g., perceptive, syntax). Furthermore, we suggest future researchers quantitatively measure the effectiveness of the generated visualizations and their interpretability to unlock the inherent black box phenomenon encompassed in a VRS (see Sec. \ref{sec:open_challenges}).

\noindent\textbf{Datasets} -- A crucial element for evaluating methods in the literature is a uniform benchmarking system that uses standardized metrics and datasets. While the other surveys mention few available datasets, they do not discuss them, highlighting problems and limitations. Contrarily, we delve deeper into the matter and classify these datasets into three main categories. We also argue, see Sec. \ref{subsec:datasets}, that the literature would benefit from open source and well-curated datasets, as pioneered by the authors in~\cite{podo2022plotly}, to push the contributions in this research field further.

\noindent\textbf{Method Taxonomy} -- To the best of our knowledge, this survey represents a pioneering effort within the expansive domain of AI4VIS by introducing a comprehensive categorization of methods found in the literature (see Sec. \ref{sec:methods} and Table \ref{tab:taxonomy}). Notably, while previous surveys such as those by Wang et al. \cite{wang2021survey} and Wu et al. \cite{wu2021ai4vis} have provided valuable insights into certain methods, they are now considered outdated, primarily concentrating on papers published before 2020. This temporal limitation results in a lack of exploration of the latest substantial advancements, such as those involving Large Language Models (LLMs). Moreover, our taxonomy is a practical guide for readers seeking the most pertinent VRSs for specific scenarios. Contrarily to Wang et al., \cite{wang2021survey}, who categorize surveyed works based on learning techniques (supervised, semi-supervised, unsupervised, and reinforcement learning), we contend that this approach lacks a principled view of the state-of-the-art. Our taxonomy goes beyond delving into the specifics of the problems addressed by the models.

On the other hand, Wu et al. \cite{wu2021ai4vis} do not offer a method taxonomy but instead discuss each method's characteristics based on criteria such as feature engineering, representation, and visualization generation mechanisms. Uniquely, our survey stands out as the first to consider the learning of \textit{constraints} as a distinctive aspect for characterizing methods. This innovative perspective aims to provide readers with a focused understanding of visualization recommenders and their inherent characteristics.

\noindent\textbf{Visualization Pipeline} -- We recognize the utility of other surveys to provide the reader with a step-by-step user-VRS interaction pipeline. These pipelines elucidate how users may generate ``effective'' visualizations in a specific context. Our survey instead treats the challenging issue of generating effective visualizations through quantitative, objective measures. Nevertheless,  we provide the reader with a pipeline showing how visualizations can be  "fine-tuned" to comply with specified constraints, as illustrated in Fig. \ref{fig:constraint_hierarchy}. To be fair to the other survey~\cite{wang2021survey}, we place a $\sim$ symbol in Table \ref{tab:survey_comparison} to indicate a partially covered aspect.

\subsection{Contributions of this Survey}
\label{contribution}
 This survey aims to contribute to a more rapid evolution of a research field that is extremely relevant from the perspective of applications and user needs, as we already highlighted in Section \ref{significance}. To this end, in addition to a discussion and comparison of published works in the domain of VRS (Sec. \ref{sec:methods}), we provide a systematic definition of the A-VRS task (Sec. \ref{sec:problem_formulation}), a summary of open challenges (Sec. \ref{sec:open_challenges}) and hints for future research directions (Sec. \ref{sec:future_work}).
 
 Furthermore, compared to previous investigations in the more general field of ML4VIS, our work compensates for some limitations and offers original perspectives, as detailed below.
 
\begin{enumerate}
    \item \textbf{User-Centered vs. Method-Centered Perspective}: One limitation of the other surveys is adopting a predominantly user-centered approach. While this is undoubtedly valuable for addressing user needs and preferences, it often fails to delve into the technical details of the underlying algorithms and methodologies that drive the visualization process. Consequently, they remain superficial in summarizing the core mechanisms that facilitate the creation of effective visualizations.
    \item \textbf{Formal Definition of the Considered Task}: We are the first to formally define and characterize A-VRSs in the context of machine learning for visualization. Unlike previous surveys, which adopt a broader point of view, we provide a structured and comprehensive definition of A-VRSs and elucidate how they operate under different constraints. Our work explicitly outlines the formal definition of A-VRSs, shedding light on their core attributes and functionality. This formal definition (see Sec. \ref{sec:problem_formulation}) serves as a foundation for understanding the role and capabilities of A-VRSs within the broader landscape of machine learning for visualization.
    \item \textbf{Lack of Micro-Level Details}: Another contribution of our survey is to provide detailed insight into how the visualization space is effectively reduced to produce meaningful, efficient, and interpretable visualizations. The other surveys present a macro-level view of visualization pipelines. Still, the intricate steps and techniques involved in reducing dimensionality and imposing constraints for enhancing interpretability are often glossed over. This deficiency limits the practical guidance provided to researchers and practitioners seeking to harness the full potential of machine learning for visualization.
    \item \textbf{Providing a Methodological Taxonomy}: The available surveys enumerate various methods from the literature without offering a systematic taxonomy or categorization of these methods. The absence of a well-structured taxonomy can make it challenging for readers to navigate the diverse landscape of machine learning techniques in visualization effectively. Organizing and classifying these methods is crucial to facilitate a more coherent and accessible understanding of the field.
    \item \textbf{Highlighting the Role of Datasets}: Notably, the other surveys largely overlook the pivotal role that well-curated datasets play in advancing the field of machine learning for visualization. Datasets serve as the foundation upon which models are trained and tested. The absence of examining the necessity and usefulness of such datasets leaves a critical gap in understanding the practical applicability and real-world relevance of the methods discussed in these surveys. Our work emphasizes the importance of high-quality datasets in driving progress in the field (see Sec. \ref{subsec:datasets}).
\end{enumerate}
\section{Problem Definition}\label{sec:problem_formulation}

Agnostic Visualization recommendation is a relatively new and potentially disruptive area of research that aims to solve the problem of transforming data into information \textit{at scale} in the current data-flooding scenario. Due to its novelty,  a precise definition of this task and its objectives has not yet been provided in the current literature.

In general terms, we can define a visual recommender as a system that takes in input a tabular dataset $\mathcal{D}$ of $n$ columns (features) and suggests a set of perceptually effective charts (visualizations), highlighting relevant and insightful feature relationships in $\mathcal{D}$.  State-of-the-art VRSs reduce the complexity of the task by providing the system with explicit knowledge of design rules and/or additional input specifications, such as the type of plots, the input variables on the plot axes, etc.  The challenging task of A-VRSs is to fully mimic the capabilities of a human analyst who attempts to discover and illustrate relevant relationships among the data variables. They can therefore be considered, at least in perspective, as a form of Artificial General Intelligence  (AGI) \cite{AGI}. 
Although a visual recommender of these capabilities is desired, most current A-VRSs try to solve a slightly simpler problem over a smaller set of features, as formalized in the following definition:

\begin{definition}\label{def:VRS_in_the_literature}
\textbf{A-VRS}: Given a tabular dataset $\mathcal{D}$ of $n$ features $\Phi = \{\phi_1,\dots,\phi_n\}$, and a chosen subset $\Phi' \subseteq \Phi$ of $k$ features in input, a VRS, based on a set of automatically induced 
%and/or manually encoded 
constraints $C = \{c_1,\dots,c_p\}$, suggests a set of visualizations $V' \subseteq V$ that provide the best insight concerning the relations among $\phi_i \in \Phi'\;\forall i \in [1,k]$ according to the constraints $c_j \in C$.
\end{definition}

\noindent It is important to note that, according to the definition provided, constraints are learned automatically during the training phase, in contrast to standard VRSs, which usually provide specifications in input and/or embed constraints into the model in the form of rules. Furthermore, at inference time, the only input allowed is a tabular dataset and the subset $\Phi'$. In what follows, we denote as \textit{A-VRSs} those visualization recommenders that adhere to the above definition and automatically learn (at least some) constraints using Machine Learning algorithms.
The task of A-VRS is typically supervised (see among others \cite{dibia2019data2vis,hu2019vizml,moritz2018formalizing}), as shown in Figure \ref{fig:workflow} in the teaser: First (Figure \ref{fig:workflow} up), the model is trained with data-visualization pairs, to learn both to identify relevant relationships between data and to visualize them in the best possible way. Next (Figure \ref{fig:workflow} down), the learned model recommends a set of possibly insightful visualizations from new datasets at inference time.

\subsection{Visualization constraints}
\label{sec:vizconstraints}
In the literature, the notions of ``constraints'' and ``insightfulness'' - or effectiveness - of a visualization lack a precise definition and are specified differently depending on the context considered. We begin with an analysis of visualization constraint types\footnote{Note that here we do not consider the problem of how to incorporate constraints into a VRS, an issue that we analyze later, in Section \ref{sec:taxonomy}}. Based on the works surveyed, we have identified four types of constraints: i.e.,  \textit{syntax}, \textit{task-based},  \textit{design}, and \textit{perceptual} constraints. Note that the following discussion defines constraints regardless of how (automatically or manually) they are incorporated into the model, so we are referring to VRSs in general.

\begin{itemize}[noitemsep,topsep=0pt]
 \item \textit{Syntax constraints} encompass a set of atomic visualization rules that conduct the correct generation of charts.  These constraints depend on the relationship between single features $\phi_i$ and the visualization type. For example, line charts require the variables on the Cartesian axes to be of type ordinal or metrics.
 \item \textit{Task-based constraints} specify whether potential visualization recommendations have (non)interesting and (non)relevant visible patterns based on user-specified objectives. These constraints affect the choice of visualization types. For example, line charts and scatterplots are appropriate for comparing feature trends. 
 \item \textit{Design constraints} are a set of aesthetic specifications that beautify the recommended visualizations to render them more memorable \cite{borkin2013makes}, like, for example, an appropriate layout, choice of colors, and scale. These constraints are applied on top of the chosen visualizations $V' \subseteq V$.
 \item \textit{Perceptual constraints} refer to the limitations of human perception and cognitive abilities that must be considered when designing a VRS. Indeed, this is a very relevant constraint since visual perception is a primary source of information as people interpret and make sense of the world around them \cite{COLE2021129}. Perceptual features depend on the relationships among features and mostly affect the insightfulness of the generated recommendations $V' \subseteq V$. Examples of perceptual features that affect visual interpretation are discrimination \cite{taylor2022visual} and constancy \cite{garrigan2008perceptual,walsh1998perceptual}. 
\end{itemize}

\begin{figure*}[!t]
    \centering
    \includegraphics[width=\textwidth]{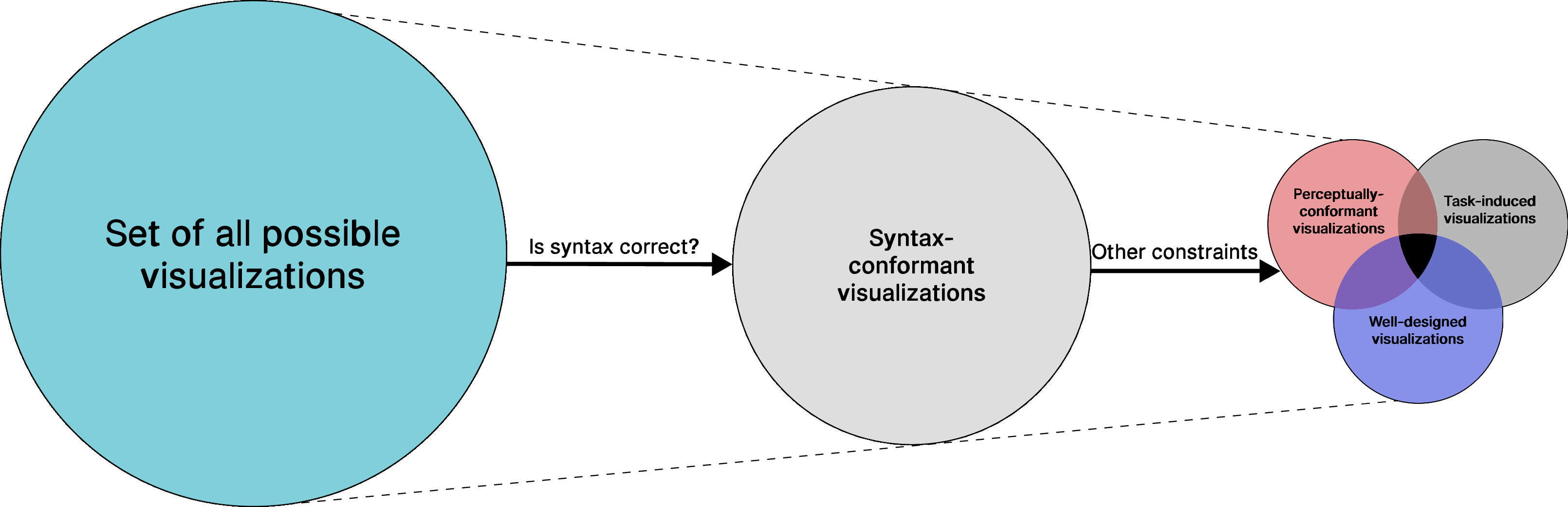}
    \caption{Application of constraints on VRS to reduce the search space of plausible visualizations. Notice that all possible visualizations from a dataset might contain representational errors rectified by inducing a principled \textit{syntax} engendering a syntax-conformant visualization subset. Once a syntax constraint is applied to the VRS, one can induce other constraints, i.e., task-, perceptual-, and design-related. These constraints can be applied in any order (hence the Venn diagram visualization). The intersection of all of them represents the visualizations with a correct syntax, a well-defined task, a perceptually conformant layout, and a well-structured design. }
    \label{fig:constraint_hierarchy}
\end{figure*}
A visualization needs to satisfy all the previous constraints to be considered a ``correct'' and adequate recommendation to the end-user, although they are not sufficient to guarantee insightfulness (a concept that we deepen in Section \ref{effectiveness}). 

Notice that the relationship between the previous constraints can be represented as a Venn Diagram (see Figure \ref{fig:constraint_hierarchy}), which permits the constraints to be applied in any order, besides those related to syntax as discussed in the following. Given a set of features $\Phi$, the syntax rules are the first set of constraints applied to the recommendation of plots. Syntactic rules are essential to avoid representational errors: e.g., if we consider a quantitative feature $\phi_i$ and a categorical/ordinal one $\phi_j$, the syntax might specify that the usage of a bar plot is advisable to convey the correct insight to the end-user, whereas, the usage of a line chart would be meaningless. Therefore, syntax constraints shrink the space of all possible visualizations into those that can be correctly rendered.

Next, the other types of constraints can be considered. Task-based constraints are optional to reduce the original features $\Phi$ to lead a VRS to focus on a subset of possible recommendations based on specific objectives (tasks), such as comparison, trend detection, etc. For example, one could make the VRS concentrate on visualization types that may help identify commonalities in specific features. In this way, the VRS, instead of exploring the space of all possible visualizations, would look for only those that could highlight the existence of clusters of data instances, such as scatterplots.

VRSs may further reduce the visualization space by selecting those most approximating particular perceptual properties. For instance, given a set of scatter plots, not all are appropriate for recommendation due to overplotting\footnote{Overplotting is when the data in a  visualization overlap, making it difficult to see individual data points.} \cite{bertini2006give}, or cluster ambiguity \cite{behrisch2018quality}, which hinder humans' ability to discriminate. Therefore, a VRS should try to select those charts that mitigate/reduce perceptual problems and best visually represent underlying feature relationships. 

Finally,  VRSs may refine the aesthetics of the recommended charts by tuning retinal properties (e.g., avoiding using indistinguishable colors) and graphical marks (e.g., legend representation), following specific design constraints. While, in principle,  the first three types of constraints can be incorporated in a VRS based on well-established sets of rules \cite{hayes1985rule}, perceptual features are by far more complex since they are mostly defined at a qualitative level \cite{heer2009sizing,reda2018graphical}. It follows that the set of perceptual constraints can be empty, thus enabling VRSs to learn the latent visual relationships of the dataset features from the charts in the training data (see \cite{dibia2019data2vis,luo2018deepeye}) and exploit the learned constraints to make recommendations at inference time. In this way,  VRSs are set to autonomously learn how to mitigate perceptual challenges, such as overplotting, while selecting the most suitable visual chart representation according to the learned design constraints.

\subsection{Effectiveness measures}
\label{effectiveness}
Considering that the visualizations that satisfy the previously described constraints are only correct from a rule-based perspective, they might not best emphasize the relationships among the features $\phi_i \in \Phi$. Therefore, they must be ordered according to a particular effectiveness/insightfulness measure.

Recall that $V$ is the space of all possible visualizations induced from the dataset features $\Phi$. According to what we discussed above, we can define a visualization generation function $f_C: \Phi \times \Phi \rightarrow V$ that, given in input a pair of features\footnote{For the sake of simplicity, we discuss only 2d charts hereafter.} $(\phi_i, \phi_j)$, provides a visualization conditioned on the constraint set $C$. Furthermore, we can define an insightfulness function $g: V \rightarrow \mathbb{R}$ that, taken in input a visualization, outputs an effectiveness score. Hence, a VRS can order the recommended visualizations $V^*$ for all feature couples $(\phi_i, \phi_j) \in \Phi \times \Phi$ as follows:
\begin{equation}\label{eq:vis_recs}
    \begin{gathered}
        v_{\phi_i,\phi_j} = \arg \max g(f_C(\phi_i,\phi_j))\\
        V^* = \text{sort}_{g(v_{\phi_i,\phi_j})} (\{v_{\phi_i,\phi_j}\; \forall (\phi_i,\phi_j) \in \Phi \times \Phi\})
    \end{gathered}
\end{equation}
According to Equation \ref{eq:vis_recs}, all feature pairs $(\phi_i,\phi_j)$ have a set of most insightful visualizations associated with them. These visualizations are then sorted according to their insightfulness score $g$ and returned to the end-user as a set $V^*$ of the best-representing plots of the underlying feature relationships, conditioned on the constraints either learned or provided in the input. Equation \ref{eq:vis_recs} guarantees to include the most insightful visualization according to function $g$ for each feature pair $(\phi_i,\phi_j)$ s.t. $\phi_i \neq \phi_j$. Finally, to alleviate the end-user from consulting many recommendations, recommender systems may return the top $\ell$ suggestions (or those with an insightfulness score above a threshold $\theta$).

Although the previous definitions of visual recommenders (see Def. \ref{def:VRS_in_the_literature}) and insightful recommendations (see Eq. \ref{eq:vis_recs}) are useful to clarify formally the problem and its boundaries, the literature on VRS has not yet proposed quantitative measures to assess  $g$ automatically. 
Memorability \cite{borkin2013makes}, emotive engagement \cite{kennedy2016engaging}, and serendipity \cite{ziarani2021serendipity} are some of the well-known measures used in the literature to assess the ``goodness'' of standard recommender systems. However, like perceptual constraints, these measures are mostly qualitative. Serendipity is a quantitative metric, whereas memorability and emotive engagement rely on crowd-sourcing (user agreement) to assess the insightfulness of the recommended visualizations. Progress in exploring effective perceptual principles was made in \cite{zeng2023review}. The authors undertook a systematic analysis of 59 academic papers, resulting in a dataset of concrete perception-driven design rules that aim to optimize the performance of VRSs.

In the absence of consolidated quantitative metrics, VRS to date rely on crowdsourcing to determine the correctness (rather than an insightfulness score) of a proposed visualization: a visualization is labeled as ``correct'' if it has been generated or approved, by human analysts (see Sec. \ref{sec:taxonomy}).

\section{A Review of VRSs}\label{sec:methods}
This section surveys the methods and solutions proposed in the VRS literature. 
\begin{itemize}
\item In Section \ref{sec:related_work}, we shortly describe the solutions adopted by the retrieved systems;

\item In Section \ref{sec:taxonomy}, we present a comparison of the state-of-the-art A-VRSs, according to several dimensions. 
\end{itemize}

\begin{figure*}[!t]
    \centering
    \includegraphics[width=\textwidth]{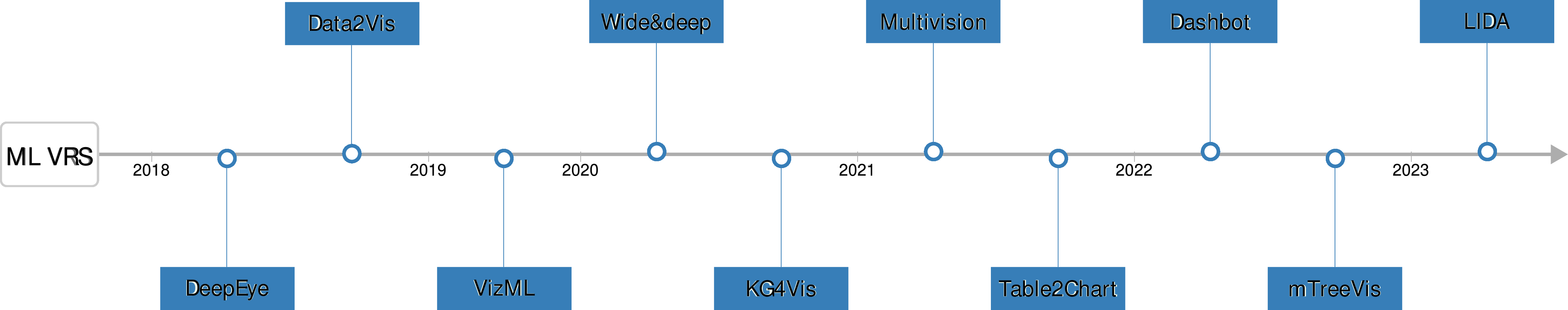}
    \caption{Timeline of A-VRSs.}
    \label{fig:vrs_timeline}
\end{figure*}

\subsection{Related work}\label{sec:related_work}
Hereafter we present a summary of papers retrieved according to the search keywords and methodology described in Section \ref{surveymethodology}. The surveyed papers are divided into \textit{constrained} and \textit{agnostic machine-learning-based}  VRSs. The former are rule-based systems and  ML-based models that allow end-users to restrict the complexity of the task by providing constraints. A-VRSs are machine-learning-based systems adhering to  Definition \ref{def:VRS_in_the_literature} provided in Section \ref{sec:problem_formulation}. Although constrained VRSs (hereafter C-VRSs) are not the main focus of this survey,  they laid the foundations of state-of-the-art A-VRSs.

\subsubsection{C-VRSs}
\label{weak}
VRSs made their first contributions in the mid-80s, with rule-based methods being the most prominent approach.
Most  VRSs are rule-based and built on the foundations of the seminal works in \cite{bertin1983semiology,cleveland1984graphical}. 
As previously stated, these systems usually allow users to specify some constraints at inference time, especially task-based ones.  A first example is APT \cite{mackinlay1986automating},  which generates, filters, and scores visualizations based on expressiveness and perceptual effectiveness criteria. Other systems, such as SAGE \cite{roth1994interactive}, BOZ \cite{casner1991task}, and Show Me \cite{mackinlay2007show} support a wider range of data, encodings, and tasks. Lastly, TaskVis \cite{shen2022visual} supports user-specified tasks - among 18 possible task choices (e.g., clustering, comparison, correlation) - that narrow down the process of recommending visualizations. Recently, hybrid systems \cite{chakrabarti2021towards,DBLP:conf/sigmod/HuOH18,viegas2018generating,wongsuphasawat2015voyager,wongsuphasawat2017voyager} have also emerged, which blend visual encoding rules with the recommendation of visualizations that include non-selected features.

Other systems use a relatively simple ML approach and restrict the possible visualization choices by allowing the end-user to specify the task. Among these, \textit{Kopol} \cite{saket2018task} provides users with a ranked list of the most effective visualization types, including Table, Line Chart, Bar Chart, Scatterplot, and Pie Chart. It employs a decision tree model trained on a vast dataset of user responses gathered through a crowd-sourced experiment. At inference time, Kopol takes in input a task (e.g., comparison, anomaly detection, etc.) and data attribute types and predicts a prioritized set of visualization types based on users' performance time, accuracy, and popularity preferences. DracoLearn \cite{moritz2018formalizing} is a system that utilizes a formal model, Draco,  to encode visualizations with hard and soft constraints, expressed as a set of rules. Hard constraints, as defined by the authors, encompass \textit{task-based and syntax} constraints that are mandatory for a visualization to be considered suitable for recommendation purposes. In contrast, soft constraints are comprised of \textit{design and perceptual} criteria that, while not essential, may still be taken into consideration during the visualization recommendation process. In DracoLearn, end-users input a partial specification that describes their incomplete intent for a desired visualization. More precisely, they specify a task and a subset of features. To answer users' queries, DracoLearn recommends visualizations that do not violate the hard constraints and are ranked highest according to the soft constraints by training a RankSVM classifier on the feature vectors of the visualizations.

Despite their effectiveness, these systems have limitations in capturing non-linear relationships and require expert or crowdsourced judgments to generate visualization rules. Furthermore, as the dimensionality of input data increases, the combinatorial nature of rules results in an exponential increase in possible recommendations unless the user is allowed to specify additional constraints, such as TaskVis and DracoLearn.

\subsubsection{A-VRSs} papers in this domain belong to the Agnostic VRSs introduced in Section \ref{sec:problem_formulation}. In these systems, at least some (if not all) of the visualization constraints and design rules are learned automatically, with the support of ML. Furthermore,  at inference time, the user only provides a dataset and, in some cases, a subset of features $\Phi'$ rather than the entire set to reduce computational complexity. Some of these systems also allow the user to refine the query after providing the first recommendations. 

Figure \ref{fig:vrs_timeline} illustrates the distribution of published papers in the A-VRSs literature in time. Hereafter we summarize the visual recommendation methodologies proposed by these systems. 

\noindent
\textbf{DeepEye} \cite{luo2018deepeye} adopts a hybrid VRS approach that tries to ameliorate, via ML strategies, rules specified by domain experts. DeepEye addresses the visualization recognition, ranking, and selection problems. The system uses a binary classifier, learning-to-rank model, experts' knowledge, graph-based techniques, and rule-based optimization to select the top-k visualizations and eliminate suboptimal ones. The authors focus on supporting the best choice of bar, line, pie, and scatter charts on real-world datasets from several domains. 

\noindent
\textbf{Data2Vis} \cite{dibia2019data2vis} is a neural translation model for automated visualization generation where tabular data is mapped to visualization specifications in Vega-Lite \cite{satyanarayan2016vega}\footnote{Vega-lite is a visualization grammar, i.e., a parametric data structure that contains all the visualization specifications. }. It relies on a two-layered bidirectional LSTM encoder-decoder architecture with an attention module and trains the model - without any pre-defined constraints - on a corpus of Vega-Lite visualization specifications with six chart types and three data transformations (i.e., aggregate, bin, and time-unit). 

\noindent
\textbf{VizML} \cite{hu2019vizml} tackles the visual recommendation problem as a procedure for making design selections to enhance effectiveness, which depends on the dataset, task, and context. The authors generate a raw dataset of one million unique dataset-visualization pairs. They elaborate on their methodology of collecting, preprocessing, and extracting features from the corpus by identifying five crucial design choices\footnote{For example, choosing how to encode a particular column along the x- or y-axis or selecting a specific chart type.} from the visualizations. Finally, they predict these design choices using a three-layer neural network with ReLU activation functions. 

\noindent
\textbf{Wide\&Deep}  \cite{qian2020ml} encodes the visualizations into meta-features based on the combination of features and some specific configuration. It is based on a ''wide" model that outputs a wide score using linear manipulations over cross-product feature transformations to capture any feature pair occurrence that leads to interesting visualizations. Next, a ''deep" model outputs a deep score using non-linear transformations to generalize unseen feature pairs that might lead to effective visualizations. 

\noindent
\textbf{KG4Vis} \cite{li2021kg4vis} bridges the gap between manual rule specifications and black-box machine learning approaches. Here, the knowledge graph models the relationship between different entities, and TransE \cite{bordes2013translating} is used to learn the embeddings of both relations and entities, thus, learning the visualization generation rules. When a new dataset is given, KG4Vis infers the relevant rules and recommends effective visualizations, offering \textit{interpretability} and automation (\textit{no domain expert knowledge for rule specification}).

\noindent
\textbf{Multivision} \cite{wu2021multivision} is a Siamese neural network approach for multiple view recommender systems to simultaneously represent different perspectives of data. It employs a bidirectional LSTM layer to predict an assessment score for data feature selections. Additionally, it incorporates a multiple-view visualization (MV), e.g. a dashboard,  as a sequence of chart embeddings using scores predicted by a single-chart assessment model and MV design guidelines \cite{DBLP:conf/avi/BaldonadoWK00}. 

\noindent
\textbf{Table2Chart} \cite{zhou2021table2charts} uses an encoder-decoder deep Q-value network (DQN) for table-to-template generation, incorporating a copying mechanism to select relevant features to fill templates. It relies on a mixed-learning approach to handle imbalanced data and trains the DQN on a multi-type task of the main chart types. Multi-type is when a ranked list of a few major types of charts (pie, histogram, etc.) are generated together for a set of variables. The authors use a shared encoder for all recommendation tasks, which is exposed to diverse source tables of several chart types. This enables the encoder to learn shared table representations containing semantic and statistical information about data features. The pre-trained table representations are then transferred to type-specific decoders for single-type tasks. Single-type tasks address the completion of visualization details for a single chart.  Table2Chart claims to be the first to address single-type tasks.

\noindent
\textbf{DashBot} \cite{deng2022dashbot} is a reinforcement learning strategy for dashboards of charts. Once presented with a dataset, Dashbot automatically explores the data and generates a ranked set of dashboards. Each dashboard explores specific subsets of key columns. The system automatically learns mutual relationships between charts to generate a unified dashboard. It can also predict chart types in light of their interrelationships, e.g. when adding a chart, chart configurations should be considered, while when changing the key column, the selected columns should be considered.

\noindent
\textbf{mTreeVRS} \cite{wang2022machine}  encompasses three layers: i.e., Subspace Importance Assessment (SIA), Visualization Type Recommendation (VTR), and Rule-based Chart Encoding (RbCE). The SIA component evaluates the importance of each attribute in the input multidimensional hierarchical tree using a random forest that selects the top three most important attributes. These are fed to the VTR component that predicts chart types for all attributes. Finally, the attribute and the predicted chart type are passed to the RbCE component, which adopts pre-defined rules to transcribe them into visualizations.

\noindent \textbf{LIDA} \cite{dibia2023lida}  introduces an innovative methodology for implementing a VRS by harnessing the capabilities of Large Language Models (LLMs)   \cite{teubner2023welcome}. This work takes advantage of the enormous potential LLMs offer to explore new frontiers in automated visualization generation.  LIDA is a modular pipeline comprising four modules, with the first three employing GPT-4 \cite{openai2023gpt}. The first module is the \textit{summarizer} responsible for extracting meaningful information from the data and then converting them into a rich text description using an LLM. Then, the \textit{goals generation} module identifies the most appropriate visualization goals based on the dataset description. Yet, the description and the goals generated are fed to the \textit{code generator} module, which generates the code for the visualizations. A final non-LLM-based module improves the design of the visualization.  To shape the output of each phase, the author applied prompt engineering techniques \cite{white2023prompt} without tampering with the model's underlying architecture. To date, LIDA is the model closest to the goal of complete agnosticism.

To facilitate comparison amongst the surveyed systems, Table \ref{tab:sota-advancement} shows, for each system, the specific problem addressed and the contribution over other methods in the literature. In Table \ref{tab:sota-advancement}, we do not aim to compare the different methodologies, which is addressed in Section \ref{sec:taxonomy}. Rather, we consider the incremental contribution of each system in terms of new capabilities added to the visual recommender, such as learning design choices, generating dashboards rather than single charts, and coping with hierarchical data.  The table clearly shows the rapid progress of systems in terms of their ability to replicate the multiple competencies of human visual analysts, with LIDA –- the latest published system –- holding the lead. It is easy to predict that, in the near future, LLM-based VRSs will get closer to the goal of fully agnostic recommenders.
% \usepackage{booktabs}
% \usepackage{graphicx}
\begin{table*}[!t]
\centering
\caption{The advancement of the state-of-the-art of A-VRSs by the surveyed works.}
\label{tab:sota-advancement}

\begin{tabular}{@{}p{0.15\linewidth}p{0.2\linewidth}p{0.55\linewidth}@{}}
\toprule
Paper &
  VRS problem addressed &
  Contribution to the state-of-the-art \\ \toprule
\textbf{DeepEye} \cite{luo2018deepeye} &
 \textit{Visualization classification (good/bad), selection and ranking }&
 One of the first systems to introduce ML techniques for the visualization recommendation task. DeepEye involves a hybrid ML \& rule-based strategy, allowing experts to specify ranking rules to support  ML models. \\ \midrule
\textbf{Data2Vis} \cite{dibia2019data2vis} &
 \textit{Visualization recommendation as language translation problem} &
  Data2Vis is the first to shape the visualization recommendation as a deep neural translation from a data specification language to a visualization specification (e.g., JSON to VegaLite). \\ \midrule 
\textbf{VizML} \cite{hu2019vizml} &
  \textit{Learn to predict design choices} &
  Unlike DeepEye, which focuses on classifying and ranking visualizations, and Data2Vis, which shapes the problem as a language translation problem, VizML focuses on a deep learning approach to predict visualization encoding choices, providing quantitative and interpretable importance measures. \\ \midrule
\textbf{Wide\&Deep} \cite{qian2020ml}  &
  \textit{Data visualization generation, scoring and ranking based on visualization effectiveness} &
   %Compared to VizML, which forces the design choices, to DeepEye, which involves a hybrid approach, and Data2vis, which solves the problem as a translation one, 
   Wide\&Deep  takes a further step in the direction of fully agnostic VRSs, by proposing an end-to-end model based on the notion of visualization configuration, a data-independent abstraction where data attributes are replaced by their general type.  
   %from the dense and sparse data attributes and design choices representation, to recommend and score visualizations.  
   \\ \midrule 
\textbf{KG4Vis} \cite{li2021kg4vis} &
  \textit{Mapping data visualization relationships into a knowledge graph} &
This paper is the only one to address explainability in VRSs explicitly. It proposes to map the data visualization relationship into a knowledge graph to improve the explainability of the recommendation process. \\ \midrule
\textbf{MultiVision} \cite{wu2021multivision} &
  \textit{Learning to recommend an analytical dashboard} &
  Rather than suggesting visual encodings for a single chart, like previous approaches, this system provides a multi-view data visualization recommendation to unveil different data perspectives. It generates multiple charts by combining single chart scoring with customized metrics.
  \\ \midrule
\textbf{Table2Chart} \cite{zhou2021table2charts} &
  \textit{Learn common patterns of chart creation – including data queries (selecting data to analyze) and design choices (how to visualize data) for multi-type and single tasks.}&
  DeepEye, Data2Vis, and VizML consider multi-type tasks where a ranked list of a few major types of charts are recommended together. Instead, Table2Chart exploits Q-learning to deal with multi-type and single tasks. Single tasks are defined as those necessary to complete a chart's details.  \\ \midrule
\textbf{DashBot} \cite{deng2022dashbot} &
  \textit{Learning to recommend an analytical dashboard} & 
  This paper is the first to propose reinforcement learning (RL) to generate a dashboard of visualizations. Table2Chart also exploits RL, but to generate single charts. Furthermore, DashBot can generate multiple data perspectives simultaneously, rather than generating multiple views indirectly by combining single views, as in Multivision. \\ \midrule
\textbf{mTreeVRS} \cite{wang2022machine} &
  \textit{Recommending visualization for hierarchical data structures} &
  Unlike the other papers focusing on navigating and generating visual insights from tabular data, mTreeVRS is a visual recommender specifically designed for hierarchical data. \\ \midrule
\textbf{LIDA} \cite{dibia2023lida} &
  \textit{Uses an LLM model to fit the agnostic VRS objective.} &
  LIDA is the first full-fledged A-VRS system based on LLMs.  It removes the dependency on subtask-specific models, is agnostic regarding the visualization language, and relies on cutting-edge LLM technology.\\ \bottomrule
\end{tabular}%

\end{table*}

\subsection{A taxonomy of A-VRSs}
\label{sec:taxonomy}

\begin{table*}[t]
\centering
\caption{Qualitative comparison of A-VRS in the literature, according to the coverage of relevant dimensions reflecting the scope and workflow. Legend for supported visualizations: B (bar), L (line), S (scatter), P (pie), H (histogram), D (dot), A (area), C (circle), T (tick), Bx (box), Hm (Heatmap), R (radio). Legend for supported interpretability: $\sim$ represents a work that supports interpretability but not end-to-end; \checkmark and $\times$ are self-explanatory.}
\label{tab:taxonomy}
\resizebox{\textwidth}{!}{%
\begin{tabular}{@{}llllll@{}}
\toprule
 & \begin{tabular}[c]{@{}l@{}}Automatically Learned \\  constraints\end{tabular} & \begin{tabular}[c]{@{}l@{}}Recommendation\\model\end{tabular} & \begin{tabular}[c]{@{}l@{}}Supported\\ interpretability\end{tabular} & \begin{tabular}[c]{@{}l@{}}Supported\\ visualizations\end{tabular} & \begin{tabular}[c]{@{}l@{}}VRS evaluation \\ metrics\end{tabular} \\ \midrule
%\rowcolor{Simple}
\textbf{DeepEye} \cite{luo2018deepeye}  & Perceptual, Task-based & \begin{tabular}[c]{@{}l@{}}DTree + LambdaMART\end{tabular} & $\sim$ & B, L, P, S & P, R, F1 \\
%\rowcolor{Simple}
%\textbf{DracoLearn} \cite{moritz2018formalizing} & Perceptual & RankSVM & $\times$ & A, B, D, H, L, S & Ranking accuracy \\
%\rowcolor{Simple}
%\textbf{Kopol} \cite{saket2018task} & Task-based, Syntax, Design & DTree & \checkmark & B, L, P, S & - \\
%\rowcolor{MLP}
\textbf{VizML} \cite{hu2019vizml} & Perceptual, Task-based, Syntax & MLP & $\times$ & B, L, S & Accuracy \\
%\rowcolor{Seq}
\textbf{Data2Vis} \cite{dibia2019data2vis}  & Perceptual, Task-based, Syntax, Design & \begin{tabular}[c]{@{}l@{}}Bi-RNN + Attention\end{tabular} & $\times$ & A, B, C, L, S, T & LSV, GSV \\
%\rowcolor{MLP}

\textbf{Wide\&Deep} \cite{qian2020ml}   & Perception, Task-based & MLP & $\times$ & All from Plotly & nDCG@k \\
%\rowcolor{KW}
\textbf{KG4Vis} \cite{li2021kg4vis} & Perceptual, Task-based, Syntax, Design & TransE-Adversarial & \checkmark & B, Bx, H, Hm, L, S & Axes accuracy, Hits@2 \\
%\rowcolor{Seq}
\textbf{MultiVision} \cite{wu2021multivision} & Perceptual, Task-based (optional), Syntax & Siamese (Bi LSTM) & $\times$ & A, B, L, P, R, S & \begin{tabular}[c]{@{}l@{}}Ranking accuracy,\\ R@k, k $\in [1,10]$\end{tabular} \\
%\rowcolor{RF}
\textbf{Table2Chart} \cite{zhou2021table2charts}  & Perceptual, Task-based & Deep Q-network & $\times$ & A, B, L, P, R S& R@k, k $\in \{1,3\}$\\
%\rowcolor{RF}
\textbf{DashBot} \cite{deng2022dashbot}   & Perception, Task-based (optional) & RL (Bi LSTM) & $\times$ & A, B, L, P, R, S & - \\
%\rowcolor{Seq}
\textbf{mTreeVRS} \cite{wang2022machine}  & Perceptual, Syntax & Bi-LSTM & $\times$ & B, L, P, S & Accuracy, P, R, F1 \\

%\rowcolor{LLM}
%\textbf{LIDA} \cite{dibia2023lida}  & Perceptual, Task-based & LLM & $\times$ & All & VER, SEVQ \\ \bottomrule
\textbf{LIDA} \cite{dibia2023lida}  & Perceptual, Task-based & LLM & $\times$ & All & VER, SEVQ \\ \bottomrule
\end{tabular}%
}
\end{table*}

To highlight the methodological differences among the surveyed recommendation approaches, we identify six comparison dimensions:
\begin{enumerate}
\item \textit{Automatic learning of constraints} illustrates which constraints, among those presented in Section \ref{sec:vizconstraints}, are learned automatically; \item \textit{Recommendation model} depicts the machine learning approach used to suggest visualizations; \item \textit{Supported interpretability} highlights whether the strategy is interpretable; \item \textit{Supported visualizations} defines which chart types are supported; \item \textit{Datasets used} describes the datasets used for model training and evaluation, and the method to create them; \item \textit{VRS evaluation metrics} depicts the evaluation measurements used in the literature. 
\end{enumerate}
In the following, we describe each dimension in detail and cover the most interesting aspects of each work (among those adhering to the definition in Section \ref{sec:problem_formulation}) according to the considered dimension. To facilitate the reader,  Tables \ref{tab:taxonomy} and  \ref{tab:datasets-papers} summarize the results of this comparison for each system and dimension. 

\subsubsection{Automatic learning of constraints}
\label{learningtasks}
Recall that we distinguish between several types of constraints (see Section \ref{sec:vizconstraints}). Essentially,  VRSs use these constraints to condition the generation of the visualizations. For each of the works in Table \ref{tab:taxonomy}, we show the constraints automatically learned during model training. Notice that all works, besides mTreeVRS, are set to learn task-based constraints automatically instead of expecting them in input from the end-user. Thus, these VRSs initially explore the search space of all possible visualization types and do not concentrate on a particular objective/goal. 
%We denote these VRSs as \textit{task agnostic}. 
More specifically, Multivision and DashBot can learn the task autonomously, but they can optionally receive the task as input constraints to shrink the initial search space of visualizations during query refinement. Unlike all the other approaches discussed, LIDA diverges from the typical paradigm of learning constraints during a training phase \cite{shen2023slimpajama}; instead, it leans on the knowledge already embedded in the pre-trained LLM model. An LLM is generally trained on a large corpus comprising diverse content, including code. Consequently, the model possesses the innate capacity to generate visualizations in code, such as Vega-Lite specifications, as discussed in \cite{maddigan2023chat2vis}. For this reason, LIDA does not change/train the model. Instead, it leverages zero-shot learning \cite{white2023prompt}  to instruct the model on how to behave and yield the desired output.

\begin{figure}[!]
    \centering
    \includegraphics[width=.8\linewidth]{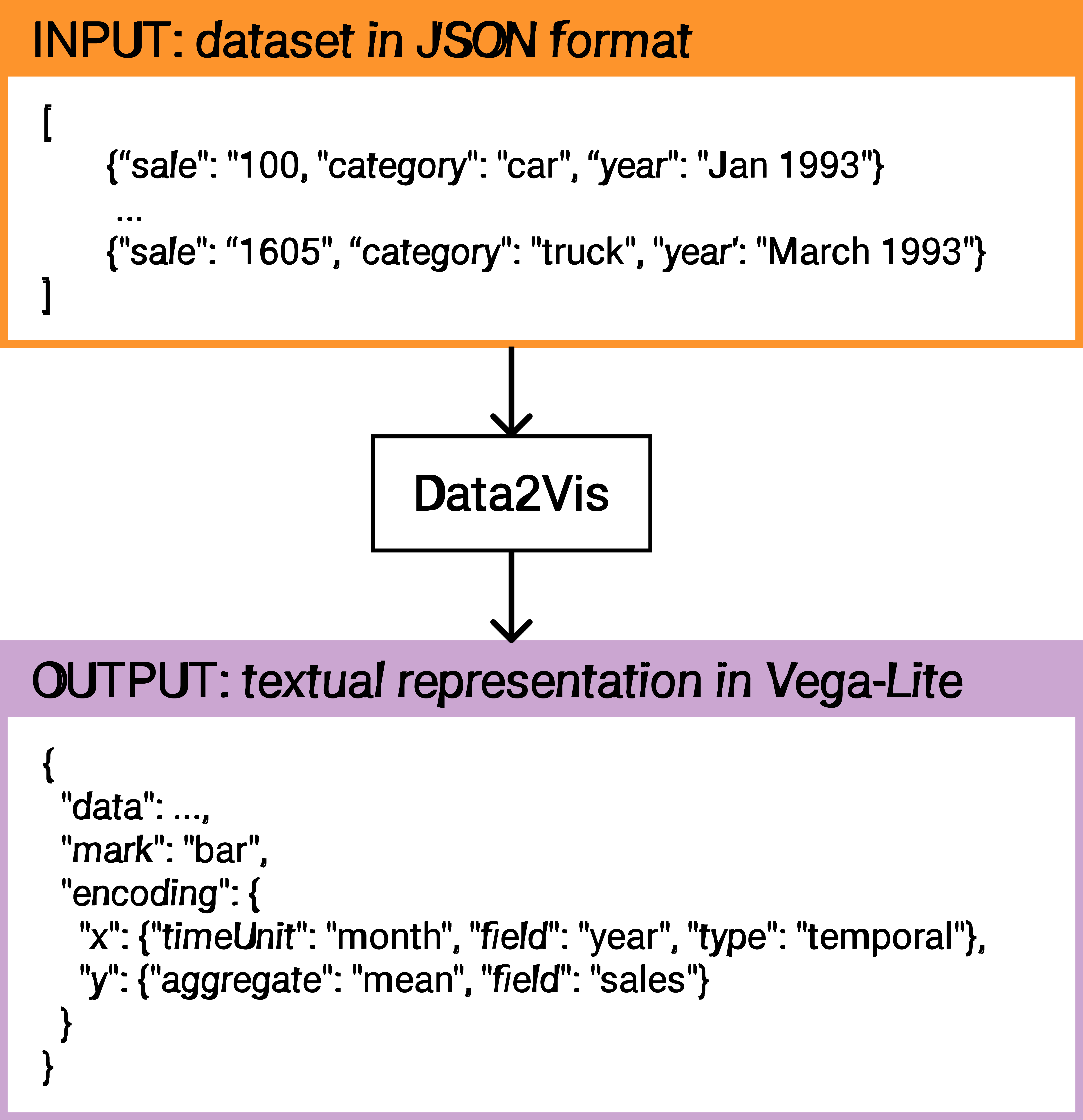}
    \caption{Example of Data2Vis translation}
    \label{fig:example_data2vis}
\end{figure}

As far as \textit{design} constraints are concerned,  the majority of works (DeepEye, %DracoLearn, 
Wide\&Deep, Multivision, Table2Chart, DashBot, mTreeVRS) rely on default and well-established constraints per chart type to make the recommendations more aesthetic rather than trying to learn them from the training visualizations and data features. Furthermore, most of the enlisted works rely on predefined \textit{syntactic} constraints that help sieve the space of all visualizations (DeepEye, Wide\&Deep, Table2Chart, DashBot). Interestingly, Data2Vis learns syntax constraints indirectly. In other words, Data2Vis treats the visualization recommendation problem as a machine translation problem -- i.e., sequence-to-sequence learning. Here, the input of the training phase depicts feature relationships, and the output is a well-defined textual representation of the desired visualization. This textual representation encodes syntax constraints. Thus, when the model learns to ``translate'' feature relationships to the textual representation of the visualizations, it inherently detects hidden syntax constraints that can be used at inference time, as shown in Figure \ref{fig:example_data2vis}. We argue that future VRS systems could adopt Data2Vis's paradigm of treating recommendations as sequence-to-sequence learning by exploiting the power of LLMs. 

Finally, as expected from ML-based VRSs, all the surveyed works can automatically learn \textit{perceptual} constraints -- although they cannot explain them -- instead of relying on domain expert rules. 

\subsubsection{Recommendation model}
It depicts how the proposed works in literature implement ML methods to suggest visualizations automatically. As shown in Table  \ref{tab:taxonomy}, these methods range from basic machine learning to advanced reinforcement learning and LLMs. For example, DeepEye uses a decision tree to determine the best chart type (scatter, line, bar, or pie) for a given set of features and then uses a LambdaMART model \cite{burges2010ranknet} to rank the charts based on effectiveness. VizML uses neural network models to make five decision choices from data, while Wide\&Deep uses neural networks to infer design decisions without constraints. Data2Vis approaches the problem as a language translation task, using a bidirectional LSTM with attention, KG4Vis uses a knowledge graph and TransE \cite{bordes2013translating} to learn embeddings on it for improving prediction interpretability, Multivision implements learning-to-rank approach using a Siamese network and a bidirectional LSTM, while Table2Chart and DashBot use reinforcement learning to generate visualizations. The former implements an encoder-decoder deep Q-value network, where the encoder learns semantic and statistical features. The latter proposes a reinforcement learning technique to generate analytical dashboards that can use well-established visualization knowledge. mTreeVRS adopts a three-layered architecture, where a random forest layer returns the top three most important attributes, a bidirectional LSTM selects the chart type, and a post-processing rule-based module translates it into the correct visualization. Lastly, LIDA exploits an LLM (GPT-4) in a four-stage pipeline as a recommendation system.

\subsubsection{Supported interpretability}
It examines the capability of visualization recommendation strategies to support end-users in understanding the reasoning behind the suggested visualizations: \textit{why was a particular visualization selected among the many possible ones?} Only a few methods provide inherent interpretability. For example, DeepEye is a method that combines a decision tree, which is interpretable, with LambdaMART, which does not provide any clear explanation of the predicted outcome, leading to the argument that this visualization recommendation strategy is uninterpretable end-to-end. In contrast, KG4Vis is the only strategy that fully supports interpretability by utilizing a knowledge graph, allowing it to illustrate the prediction steps by traversing the graph from the initial concept to the final prediction outcome (concept leaves). Due to the complexity of interpreting deep learning \cite{fazi2021beyond,olah2018building}, the remaining studies do not consider interpretability as an integral part of their strategy. For example, although they capture the latent perceptual properties of charts during the training phase and use them to select the best charts at inference time, they cannot explain these features.

\subsubsection{Supported visualizations}
It defines which chart types are supported by the systems. Notice that here we are examining the case where the chart types are not specified as task-based or design constraints, as for C-VRSs; rather, they represent an inherent constraint rising from a limitation (bias) of the visual representation types available in the provided Training set. Since models are trained on a few chart types, they can only learn to predict the best visualizations of the observed types. The most recurrent chart types are scatter, line, and bar plots. Wide\&Deep is the only method that considers all the visualization types available in the Plotly repository. Additionally, KG4Vis supports two less frequently used chart types (i.e., box plots and heatmaps). Although this aspect is not specifically analyzed in the related paper, our intuition is that LIDA can rely on a larger visualization space belonging to GPT-4 knowledge thanks to its LLM-centric underlying structure. 

\subsubsection{Datasets}
\label{datasets}
A VRS dataset is structured as a collection of data-visualization (hereafter data-vis) pairs. Starting from available datasets, visualizations are generated exploiting some visualization grammar or language, for example, using a  JSON-like syntax, which defines the structure and appearance of the plot, or the high-level Vega-Lite syntax, designed for creating interactive visualizations. Regardless of the syntax used to generate visualizations, the plotted variables and types of charts are selected manually, either through crowdsourcing or self-crafted by the VRS authors and their research groups, except Data2Vis, where a rule-based approach is adopted. Furthermore, not all authors make their data available, as shown in Table \ref{tab:datasets-papers}.

In  Data2Vis, the visualizations are obtained from 11 datasets using a set of heuristics designed to generate plots based on data and perceptual features. In this way, they generate an initial set of 4.3k Vega-Lite visualizations that are subsequently expanded using a repetitive sampling strategy, resulting in 215k  pairs with six distinct visualization types (area, bar, circle, line, point, tick)  and three different transformations (aggregate, bin, time-unit).  VizML, Wide\&Deep, KG4Vis,  and mTreeVRS use a crowd-sourced dataset, Plotly, a collection of 2.3 million data-vis pairs created by 140k users on the Plotly Community Feed.  The authors of Table2Chart and Multivision use a manually crafted dataset of 200k pairs for training and Plotly for evaluation. The authors of DeepEye do not exploit data-vis pairs for training. Rather, they use visualization pairs with annotations to identify the best chart in a pair. Similarly to the previous, LIDA does not involve any training dataset, as it leverages the inherent visualization knowledge embedded in the LLM, eliminating the dependency on external datasets. For evaluation purposes, LIDA exclusively utilizes the Vega-datasets\footnote{https://github.com/vega/vega-datasets} resource.
In DeepEye, all possible pairs of visualizations are generated from 42 datasets and subsequently annotated by 100 students. Unfortunately, the dataset is not available. An interesting observation is VizNet \cite{viznet}, a large-scale corpus exceeding 31 million datasets compiled from diverse sources such as open data repositories, visualization galleries, and online visualization tasks. This extensive dataset, the largest to date, facilitates the training and evaluation of ML models for data analysis and visualization recommendation.
\begin{table}[!t]
\centering
\caption{The datasets used in the literature. We report also the process employed to generate their data-vis pairs (when applicable). Legend: 1 - \url{https://github.com/mitmedialab/vizml}; 2 - \url{https://github.com/victordibia/data2vis}; 3 - \url{https://github.com/uwdata/draco}; 4 - \url{https://viznet.media.mit.edu/}. \textit{not applicable}  refers to a learning-to-rank procedure that requires pairs of visualizations annotated as good/bad, rather than data-vis pairs. The ''-" sign refers to either the absence of a link to the adopted dataset or to the absence of a data generation process. }
\label{tab:datasets-papers}
\resizebox{\linewidth}{!}{%
\begin{tabular}{@{}llll@{}}
\toprule
Work &
  Generation process & Data-vis pairs &
  Link \\ \midrule
DeepEye                  & Crowdsourcing (100 users) & \textit{not applicable}    & - \\
\begin{tabular}[c]{@{}l@{}}VizML, Wide\&Deep,\\ KG4Vis, mTreeVis\end{tabular} &
  Crowdsourcing (140k users) &
  2.3m &
  1 \\
Data2Vis                 & Rule-based                & 215k & 2             \\
Table2Chart, Multivision & Manually-generated        & 200k & - \\
%DracoLearn               & Internally created        & \textit{not applicable}  & 3  \\
LIDA               & -        & \textit{not applicable}  & -  \\
VizNet & Crowdsourcing + Web Crawling & \textit{not applicable} & 4  
\\ \bottomrule
\end{tabular}%
}
\end{table}

\subsubsection{Evaluation metrics}
We first note that, as anticipated in Section \ref{sec:problem_formulation},  all VRSs evaluate the performance only concerning the (binary) ground truth represented by annotated data rather than providing some measurable notion of perceptual insightfulness. "Good" visualizations are only those generated by some user concerning a given test set of data-viz pairs.  Consequently, all works presented in the literature adopt standard evaluation metrics for recommender systems\footnote{We refer the reader to \cite{bobadilla2013recommender} for a review of general-purpose recommender systems and their evaluation, as this is out of scope in this survey.}: i.e., Recall@k (used in Multivision and Table2Chart), and  Hits@k (used in KG4Vis). However, VizML, DeepEye, and mTreeVRS use evaluation metrics such as accuracy, precision, recall, and F1 without considering the ranking aspect of the suggested visualizations, leading to an incomplete understanding of the model performances. Additionally, Multivision relies on ranked accuracy to account for the imbalance phenomenon. Differently, Data2Vis uses custom metrics such as Language Syntax Validity (LSV) and Grammar Syntax Validity (GSV) to address the lack of appropriate metrics for sequence-to-sequence models in evaluating data transformation into visualizations. KG4Vis, besides relying on Hits@k \cite{bordes2013translating} for evaluating the performances of their VRS, uses accuracy to determine whether the features used in a particular recommended visualization belong to either one of the axes (binary classification). In a distinctive approach, LIDA involves custom metrics such as Visualization Error Rate (VER) and Self-Evaluated Visualization Quality (SEVQ) to address the challenge of evaluating LLM-generated visualizations. VER is intended to measure the percentage of visualizations that can be correctly compiled; differently, SEVQ is involved to measure the quality of the visualization. 
Finally, Wide\&Deep relies on nDCG@k, a modified version of the classic nDCG metric, to account for a correct evaluation of the relevance of the suggested visualizations in VRS. 
\section{Open Challenges}
\label{sec:open_challenges}
Implementing A-VRSs presents challenges that require more robust and effective systems. We identified five recurrent and critical challenges currently hindering the advancement of this potentially disruptive research field in the hope that this may guide future studies.

\subsection{Assessing the effectiveness of visual recommendations} 
According to the survey presented in Section \ref{sec:related_work}, all systems evaluate the performance of the proposed method w.r.t. a "ground truth" represented by a dataset of human or rule-based visualizations generated from a dataset $\mathcal{D}$. This is a relevant limitation, since:
\begin{enumerate}
\item the effectiveness $g$ of suggested charts (as defined in  Section \ref{effectiveness}) depends on several  factors affecting their quality, like the  context, the user's experience, and the insight the user is looking for;
\item perceptual metrics should be incorporated in $g$, but quantitative metrics are not available;  
\item the available datasets on which trained models can be biased or riddled with inexpressive charts entered by non-expert users, leading to a biased estimate of $g$. 
\end{enumerate}

Concerning the first aspect, the effectiveness $g$ of a proposed set of visualizations could be assessed according to the following factors, as discussed in \cite{luo2021empowering} (among others): 
\begin{itemize}
	\item \textbf{data accuracy}: it measures how well the model performs in retrieving an ''informative"  subset $\Phi'$  of features to show in a visualization (see Def. \ref{def:VRS_in_the_literature}) from the input dataset.
	\item \textbf{mark accuracy}: it returns the accuracy of a model in choosing the best chart type (e.g., histogram vs. scatterplot, etc.) according to design and perceptual constraints. For example, scatter plots that show the densest areas (clusters) or some visible trend capture the attention since they tell us something relevant about the relationship between plotted variables. 
	\item \textbf{axes accuracy}: it measures the accuracy of the model in placing the selected columns in the most relevant axes;
	\item \textbf{data transformations accuracy}: it returns the ability of a model to pick the most suitable data transformation function on the plot axes, e.g., sort, sum, average, etc.
\end{itemize}
%\textcolor{red}{PV fin qui}

We have already remarked on the difficulty of defining quantitative measures regarding perceptual metrics. Amongst the few studies in this area, Berish et al. \cite{behrisch2018quality} are working on creating cross-chart perceptual metrics for VRS. 

Finally, we note that metrics also depend on the consistency of the benchmark (the ground-truth dataset used to train and test a model). If the benchmark is biased, for example, if it only includes a subset of possible chart types or includes low-quality examples entered by non-expert users, the accuracy of the learned model can be over or underestimated.  This issue is deepened in Section \ref{subsec:datasets}.

%VRS interpretability - previous title
\subsection{Unlocking the black box and the interpretability of VRS}
A-VRSs rely on deep learning and have performed better than other model types in generating significant recommendations. However, they suffer from the so-called "black-box" problem, which hinders interpreting what happens "under the hood". Thus, interpretability is an important aspect of developing and deploying these systems. According to the European Commission \cite{european2020artificial}, interpretable systems create safer digital environments for users, developers, and business stakeholders, while encouraging privacy, trustworthiness, and fairness.  In critical scenarios - e.g., healthcare and finance - the lack of interpretability makes it difficult to trust the model.  Moreover, visualizations could be misleading if the model's decision process cannot be unrolled and interpreted. Imagine a non-expert user utilizing an uninterpretable VRS to suggest charts for data on non-smoker patients with lung cancer. The VRS suggests a chart showing the relationship between mutations of certain genes \cite{dabir2014ret} and the likelihood having lung cancer. A trained doctor would know that lung cancer is not only caused by these mutations, but a less experienced user might misunderstand the chart. To prevent this, VRSs deployed in critical scenarios (like healthcare, finance, and many others) must incorporate strategies that inherently provide the end-user with human-readable cues that motivate the chart suggestions. Although state-of-the-art explainability mechanisms designed for image classification (e.g., \cite{imageinterpretability} among others) might be a useful starting point, they are not readily useful to indicate which parts of a chart convey relevant and perceptually visible information on an existing relationship between plotted variables. Note that this issue is closely related to the problem of finding good perceptual metrics.

\subsection{Navigating the sparse search space in visualization generation}
\label{complexity}
Although fully-fledged A-VRS should be task-agnostic\footnote{In other words, they should be able to suggest insightful visualizations for each possible relevant task.}, we observe that they face dimensionality challenges regarding the number of axes, the number of charts, and the number of suitable feature combinations. For instance, the number of possible chart suggestions in 2d is $\mathcal{O}(\rho\cdot\binom{|\Phi|}{2})$, representing the combinations of features $\phi_i \in \Phi$ without repetitions for $\rho$ different chart types. For the sake of the argument, let us suppose that all the $\mathcal{O}(\rho\cdot\binom{|\Phi|}{2})$ 2d chart combinations are valid and are recommended to the end-user\footnote{We are aware that the entire search space of visualizations is cut off according to some significance threshold which is generally a learned function in ML-based VRS.}, and we suggest only bar plots (i.e., $\rho=1$). While the problem is tractable when $\Phi$ is relatively small or the user selects a smaller $\Phi'$, large feature spaces would entail many potentially significant charts to examine. For example, when $|\Phi|=10$, the end-user needs to assess 45 different bar plots to determine whether the information conveyed is useful. 

Notice that the search space for visual recommendations can be complex even if the VRS incorporates task-based constraints. In these cases, the overall 2d chart combinations are bounded/reduced according to the number of features requested for analysis by the end-user instead of searching in the entire feature space $\Phi$.

\subsection{Obtaining high-quality datasets}
\label{subsec:datasets}

Very few datasets have been made available to the research community, and their quality is not high. 

In  Data2Vis, the quality of data-visualization pairs is tied to the initial rules used to generate them, potentially leading to bias in the entire dataset. 
VizML, Wide\&Deep, KG4Vis,  and mTreeVRS use the Plotly crowdsourced dataset. The Plotly corpus, besides charts generated by experts,  contains bogus plots generated by non-experts which pollute the overall visualization distribution, rendering it biased towards insignificant charts. To mitigate this issue, Podo et al. \cite{podo2022plotly} present a cleaned version of the Plotly dataset. It is obtained by disambiguating insightful charts from non-insightful ones using a hybrid ML-and-human-in-the-loop approach to reduce bias and increase the dataset's quality. Although the proposed approach is promising, the need for a manual filtering step reduces its scalability to large, noisy datasets.  In conclusion, the performance of A-VRSs is hindered by the quality of the datasets employed for training and evaluation. The use of biased or noisy datasets may lead to recommendations that exclude certain groups of visualizations, and small datasets can limit the diversity and generalizability of the recommendations. Addressing these limitations is crucial for advancing the applicability of VRSs in real-world scenarios.

\subsection{Task-agnostic vs. task-aware}
\label{task-aware}
For a fully-fledged A-VRS, the ambitious goal is to generate a visualization from a dataset without the human contribution to the task definition. Recall that task-based agnosticism (see Section \ref{learningtasks}) means that the end-user receives the best chart types that encode hidden insights from data without pre-defined constraints. Although all surveyed systems tried to achieve this goal (see Section \ref{sec:taxonomy}), they are inherently limited by the types of charts in the training set.  Furthermore, as we observed in Section \ref{complexity}, navigating the full space of solutions might be intractable. Therefore, we argue that a task-aware VRS might still be useful in critical domains to guide the search mechanism toward useful visualizations w.r.t. a user-specified task. Some rule-based methods surveyed in Section \ref{sec:related_work} allow the end user to specify a task. While this makes the VRS problem more tractable, it also requires that users have sufficient experience in the types of visualizations and information they might convey (e.g., funnel charts to trace progress towards some objective), an assumption that strongly limits the prospective users of a visual recommender.
A better and relatively unexplored solution (at least concerning the objective of A-VRS) would be to let the user define contexts and tasks through lightweight specifications, e.g., in natural language. This would considerably alleviate the competence expectations of system users. NL2VIS (natural language to visualization)  is one of the most promising and recent research patterns in ML4VIS. It involves developing methods for interpreting natural language queries (NLQ) and translating them into visualizations that accurately represent the underlying data. They are constrained task-based VRSs in which the task is naturally specified, e.g., a sentence. Note that these systems do not adhere to our definition of A-VRSs in Section \ref{sec:problem_formulation}. NL2VIS  is less ambitious than  VRSs since it implies that the user has some insight, or at least curiosity, on possible existing relations among the features $\Phi \in \mathcal{D}$. On the other hand, they are still interesting since they greatly facilitate data exploration by non-expert users. 

However, performances of these systems typically worsen with lexically and semantically complex input queries, for example: \textit{What are the number of start date of the apartment bookings made by female guests (gender code Female) for each year? Plot a bar chart}. 

Developing new NLP approaches based on deep learning has recently led to new opportunities in this area. Technologies such as the LLMs let free text be managed and translated without the constraint of well-written NL queries, resulting in an essential advancement in this field.

To address interested readers, we summarize this area's most relevant and recent works. 
In \cite{chen2022nl2interface}, the authors propose a novel method for translating natural language queries into visualization using Codex - a pre-trained deep learning model developed by OpenAI for natural language to code conversion \footnote{\url{https://openai.com/blog/openai-codex}}.
They feed the LLM with natural language-SQL pair examples and the user's natural query to help the model understand the task. Then the result is converted into Vega-Lite specifications using a rule-based approach \cite{chen2022pi2}.
Notably, the authors demonstrate that their model can achieve valuable results without any model tuning or edits, highlighting the crucial role that LLM models can play in the field.
Similar to the previous method, \cite{maddigan2023chat2vis} proposes to use an LLM with zero-shot learning by exploiting prompt engineering\footnote{\url{https://en.wikipedia.org/wiki/Prompt_engineering} } techniques. They define a prompt template to translate the natural language query into a fully functional Python code to represent data in a visualization\footnote{E.g., \url{https://matplotlib.org/}}.

\section{Summary of Future Research Directions}\label{sec:future_work}
Despite their great potential in the field, state-of-the-art A-VRSs are often impractical for real-world scenarios due to the open challenges outlined in Section \ref{sec:open_challenges}. 
One major challenge facing VRS is more interpretability for deep learning strategies. In critical fields, the absence of interpretability negatively affects the usability of VRS, making them virtually unusable. As such, an area of potential investigation for future studies would be the creation of more \textit{sophisticated and user-friendly built-in interpretability techniques} aimed at assisting end-users in comprehending the visual suggestion mechanism. Additionally, interpretability can be enhanced by incorporating domain-specific knowledge into the VRS, which can aid in identifying meaningful insights in a specific context and effectively reduce the visualization search space.
Another significant challenge facing VRS is the lack of standardized and effective quality metrics. In particular, researchers should concentrate on defining quantitative metrics that may capture the perceptual insightfulness of visualizations.  This limitation is further exacerbated by the diversity of constraints (see Section \ref{sec:problem_formulation}) in different application contexts.  Furthermore, having these metrics would allow researchers to improve existing datasets by reducing bias and discarding non-insightful visualizations, and could even pave the way for unsupervised approaches that are totally absent in the literature\footnote{We emphasize that LLM-based methods rely on the knowledge embedded in the pre-trained language model. Usually, these systems are subjected to a fine-tuning phase in zero or few-shot learning, but this does not change their supervised nature.}.
Finally, to make VRS usable in real-world and high-stakes situations, future research could focus on LLM task-aware VRSs, rather than one-shot and task-agnostic recommendation approaches, which seem too ambitious in real scenarios. This suggests focusing on approaches to generate insightful and effective visualizations for a provided dataset (e.g., \cite{DBLP:journals/corr/abs-2204-06125,DBLP:conf/icml/NicholDRSMMSC22}) given lightweight task-based constraints,  for example,  expressed with an input query in natural language.
\section{Conclusion}\label{sec:conclusion}

The need for VRSs is a consequence of the proliferation of daily data and a shortage of data analysts, which has led to growing interest in the field. To the best of our knowledge, this is the first systematic survey on Agnostic VRSs (A-VRSs) are a type of visual recommenders with the ambitious goal of learning to mimic human visual analysts in discovering insightful visualizations from data without the help of human-provided constraints and/or rules. In our study, we aimed to provide a comprehensive overview of the current state of these systems and their potential for real-world applications.

We began by thoroughly reviewing the literature to establish a uniform definition of A-VRSs. Then, we analyzed the specific constraints and effectiveness measurements central to A-VRSs, highlighting the importance of factors such as interpretability, the ability to learn multiple constraint types, and the definition of appropriate evaluation metrics to rank recommended visualizations.  Our analysis identified several open challenges that may impede the widespread adoption of A-VRSs in real-world complex scenarios. Furthermore, we also guided future research to address these challenges and further advance the field.

\bibliographystyle{IEEEtran}
\bibliography{bibliography}


\vfill

\end{document}